\documentclass{article}
\usepackage{spconf,amsmath,graphicx}
\graphicspath{ {images/} }

\title{II-FCN for skin lesion analysis towards melanoma detection}
\name{Hongdiao Wen}
\address{School of Electronic Engineering/Center for Robotics\\
University of Electronic Science and Technology of China(UESTC),Chengdu,China}
\begin{document}
%
\maketitle
\begin{abstract}
Dermoscopy image detection stays a tough task due to the weak distinguishable property of the object.Although the deep convolution neural network signifigantly boosted the performance on prevelance computer vision tasks in recent years,there remains a room to explore more robust and precise models to the problem of low contrast image segmentation.Towards the challenge of Lesion Segmentation in ISBI 2017,we built a symmetrical identity inception fully convolution network which is based on only 10 reversible inception blocks,every block composed of four convolution branches with combination of different layer depth and kernel size to extract sundry semantic features.Then we proposed an approximate loss function for jaccard index metrics to train our model.To overcome the drawbacks of traditional convolution,we adopted the dilation convolution and conditional random field method to rectify our segmentation.We also introduced  multiple ways to prevent the problem of overfitting.The experimental results shows that our model achived jaccard index of 0.82 and kept learning from epoch to epoch.
\end{abstract}
\section{Introduction}
\label{sec:intro}
\begin{figure}[htb]

\begin{minipage}[htb]{1.0\linewidth}
  \centering
  \centerline{\includegraphics[width=8.5cm,height=8cm]{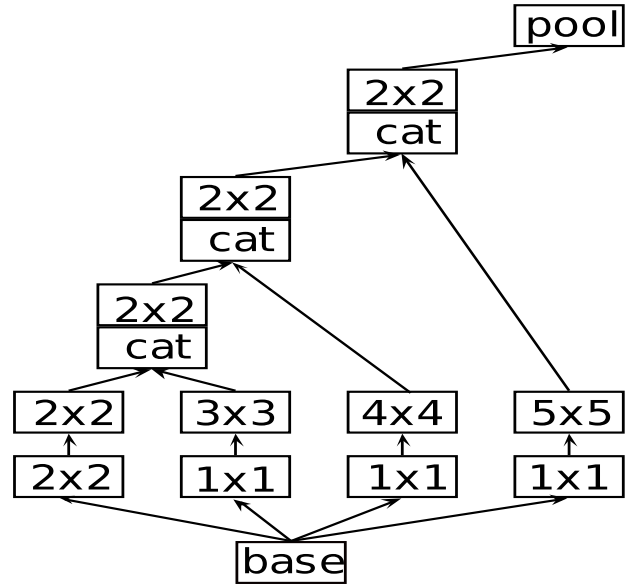}}
  \caption{The identity inception block in our II-FCN model.The base block means the input image or the output of previous block.The blocks in the middle shows the combination of layer depth and kernel size of our model.The different branches will be concatenated together,then 2x2 convolution is designed to add non-linearty and map the size of feature map equal to the next branch.Max pooling is conducted at last.The identity inception block can be reversed as block of convolution transpose.}
   \label{fig:model}
\end{minipage}
\end{figure}
Melanoma, also known as malignant melanoma, is a type of cancer that develops from the pigment-containing cells known as melanocytes.Melanomas typically occur in the skin but may rarely occur in the mouth, intestines, or eye.In women they most commonly occur on the legs, while in men they are most common on the back.Sometimes they develop from a mole with concerning changes including an increase in size, irregular edges, change in color, itchiness, or skin breakdown.Melanoma is the most dangerous type of skin cancer. Globally, in 2012, it occurred in 232,000 people and resulted in 55,000 deaths\cite{melanomawiki}. \par
Dermoscopy image detection is helpful to assist the diagnose and prevention of skin lession like melanoma.However,the melanoma in dermoscopy image is diversary in color,contrast,and shape,even human beings would have a hard time to recognize.To improve the accuracy and efficiency in melanoma detection,automaticly segmentation of melanoma is highly evaluated for medical usage.\par
Some of former efforts for segmentation is based on hand-crafted features\cite{3Dwavelet,leverage}.These hand-crafted features methods are limited and unstable due to the weak non-linear representation ability.In recent years,many works torward object segmentation like melanoma has been based on deep convolution network\cite{segdcnn}.Benefiting from the excellent ability to extract abstract features from original image,the model based on dcnn is more likely to work well in melanoma segmentation.However,more deep the network,more large should the dataset,otherwise it would take the risk of overfitting. \par
He, Kaiming, et al\cite{residual}proposed the residual network which overcomed overfitting problems of deep network.The residual scheme should be deep enough to achive complicated non-linearty.The shorcut in residual block is to carry message of features at lower layer to higher layer,if the number of residual blocks is less,then the output would be much influenced by features in lower layer.While many segmentation model are based on fully covolution networks\cite{FCN}which made up of encoder and decoder,this makes the residual block too heavy for segmentation.Meanwhile,the inception block\cite{inception} intergrated with different layer depths and kernel sizes,which is also able to convey diversity semantics to next block.In this paper, we formed a identity inception block,which can fuse different convolution branches together without any zero padding.Our model is basicly made up of 10 this kind of blocks,the total size of our model is comparablely smaller.\par
In order to train a deep model with millions of parameters,there exists many  models took image augmentation as a basic way to overcome overfitting\cite{vgg,inception}.We specified our dynamic dataset augmentation method in detail in part \(2.1\),this made our model feed with new image content from epoch to epoch.Besides,a loss function normalization was adopted to balance the sample number of pixels.Then,we proposed a approximate loss function for jaccard index metrics in addition.\par
As described in \cite{dilation},the dilation convolution is best suit for semantic segmentation benefiting from the ability to cover more large reciptive field .In our model,we add several dilation convolution at last to fill the black holes.Then we adopted the conditionnal random field\cite{textonboost,crfseg,cfrcode} to drop the small objects and rectify the boundary.\par
In this paper,we propose a original model named II-FCN with several training methods and conditional random field technique to segmentate the melanoma from dermoscopy image.Meanwhile,we introduced a identity inception block which can be reversible to build a symmetrical FCN.We proposed an loss function for better training our model. Note,our model size is more small so as to bring a light solution compared with formally FCRN model\cite{fcrn}.The contributions of our work can be summarized in the following:\par
(1)We proposed a identity inception block with four valid padding convolution branches combined the ability to extract fussion abstract features and stable the training process.Different from residual block,our implement of identity inception block enjoying non-linear feature learning without identity mapping and can be applied with the FCN model with less layer.\par
(2)Besides.We Embedded the dilation convolution at the last of our II-FCN network,which is designed to remove the black holes in object.In order to adjust the boundary according to the texture of original image,we applied the conditional random field as a post processing method to generate the final result.\par
(3)We also introduced several dataset augmentation policies and trainning tricks with our II-FCN model to improve the generalization of our model.The experiments shows that this strategy is helpful to reduce the difference of data distribution between trainning set and test set.we believe this is generasively effective to many other sematic segmentation tasks.\par
The remainning content of this paper is organised as follows.We introduced our methods in Section 2.Exeperiments and results in Section 3.Concludes at Section 4.
\section{Methods}
\begin{figure*}[htb]
\includegraphics[width=3.5cm, height=4.7cm]{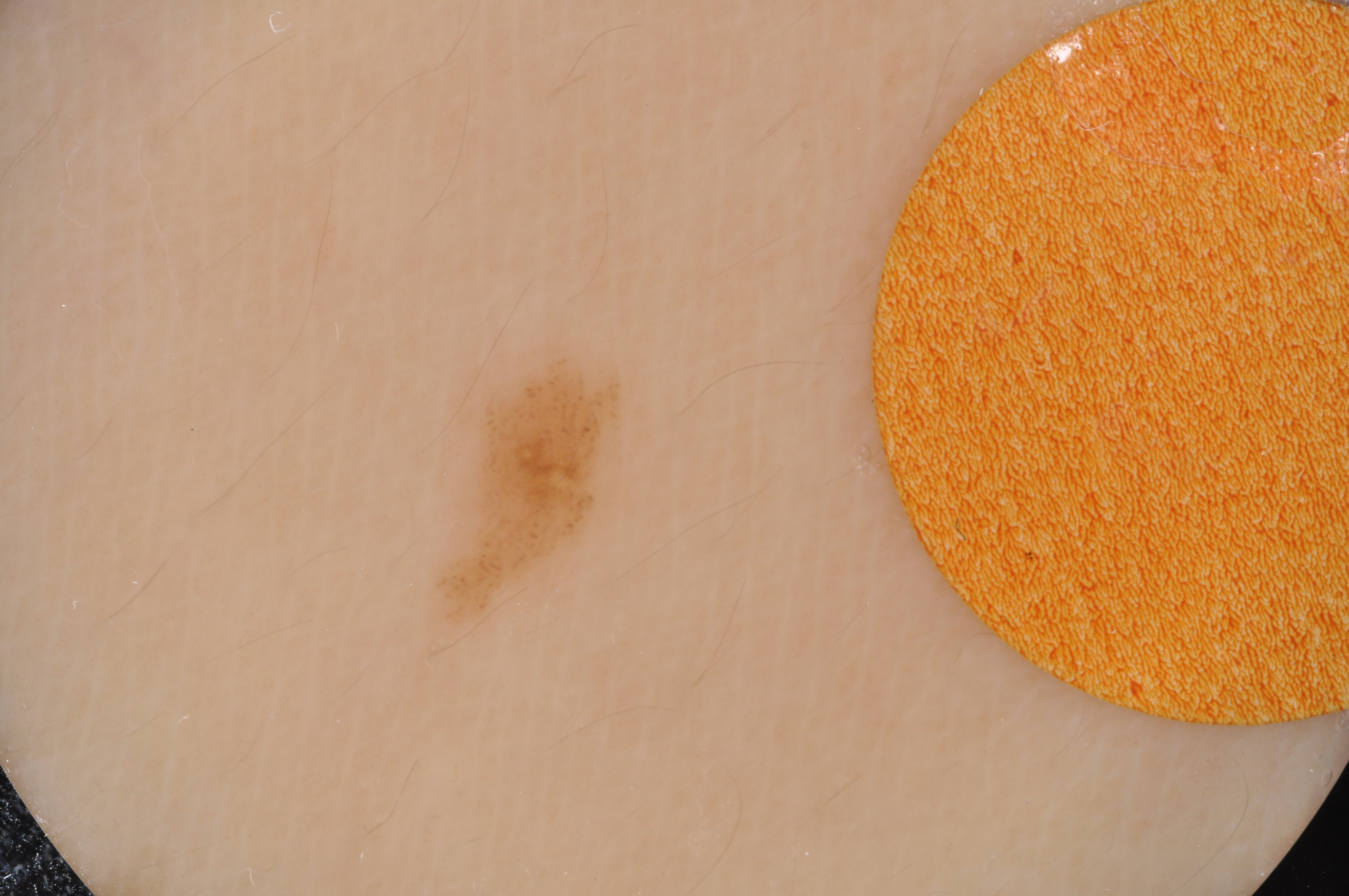}
{\includegraphics[width=10.5cm,height=4.7cm]{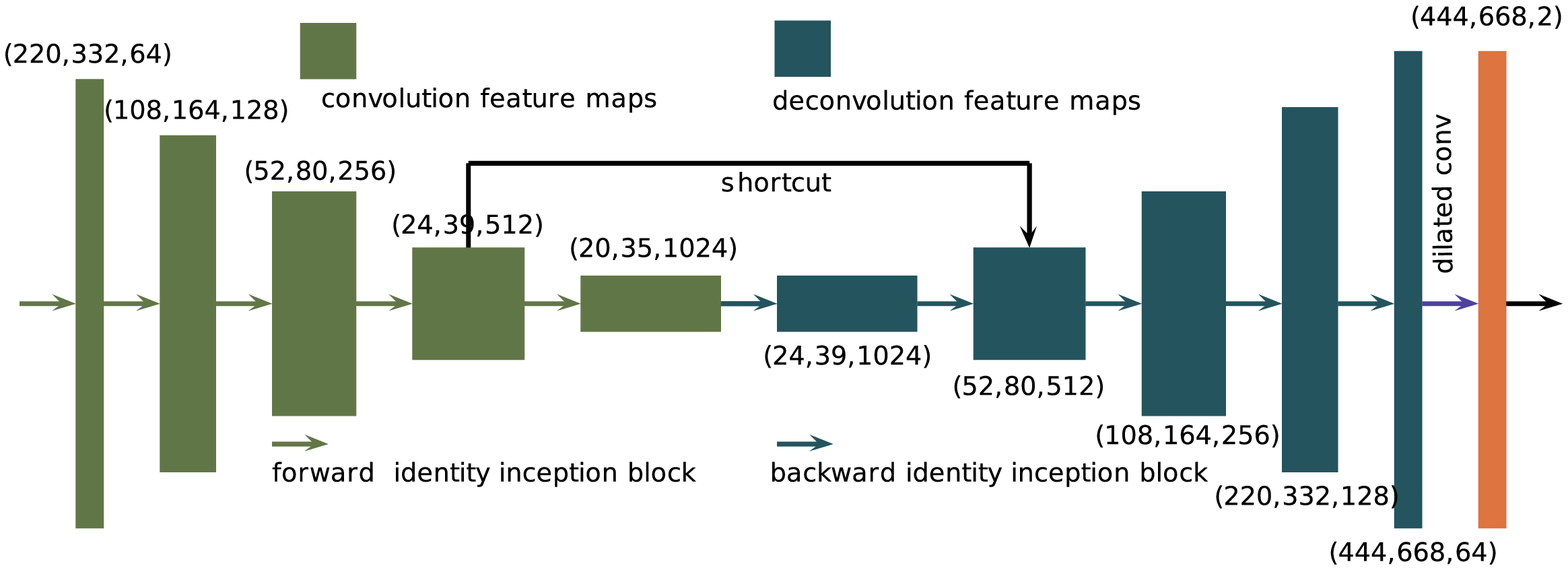}}
\includegraphics[width=3.5cm, height=4.7cm]{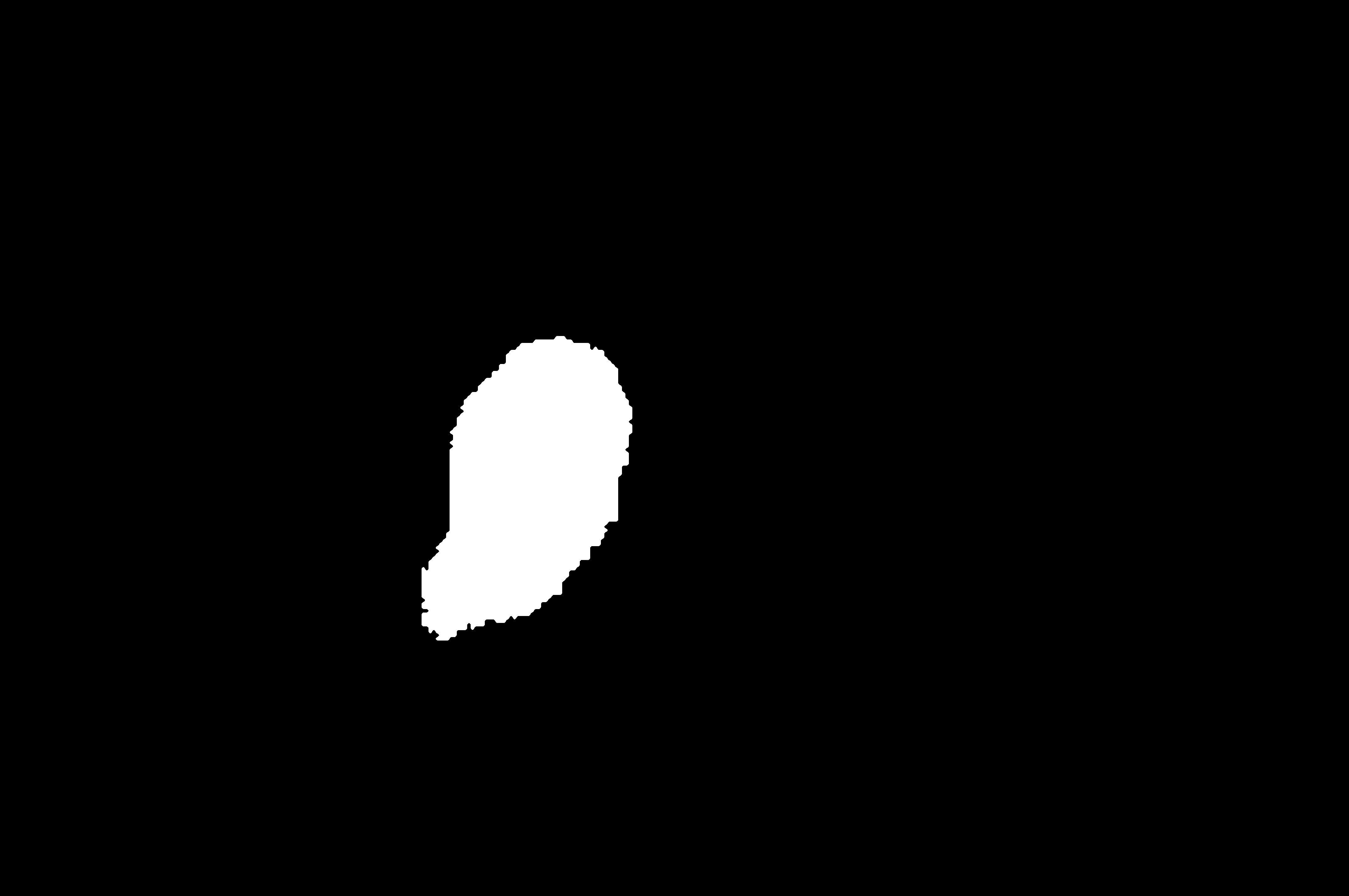}
  \caption{We constructed 5 forward identity inception block and 5 backward identity inception block in our II-FCN segmentation model.The shortcut took a residual scheme to keep texture deatails.The dilated convolution at last is to drop black holes.The sample segmentation is taken from validation set,best viewed on the color sensitiveness.}
   \label{fig:model}
\end{figure*}
\subsection{Dynamic Dataset Augmentation}
As introduced above,due to the weak objectiveness of melanoma and nonobvious　boundary between melanoma and the background,the melanoma segmentation images are difficult to label.The dataset in this challenge contains only 2 thousand training images,the distribution of training set and test set is highly different.We thought that dataset augmentation is essential to train a model with millions of parameters.We specifies our data augmentation strategies as follows.\par
(1)First,the original image has possibility of 0.2 to adjust contrast,color,apply gaussian blur and histogram equalization respectively.\par
(2)Then,the processed image from (1) has 0.2 possibility to flip left and right,flip up and down and rotate a random degree respectively.\par
(3)Last,the processed image from (2) has possibility of 0.5 to zoom in a random area.\par
The corresponding segmentgation performed the same operation as original im age.To conclude,our dataset augmentation method is able to generate different state of original image from epoch to epoch.
\subsection{II-FCN Model}
In order to build a strong and robust model to better extract features from original image,we propose the Identity Inception block which is able to combine branches with different layer depth and kernel size to generate equal sized feature maps without zero padding.The Identity Inception block can be taken as a basic feature map transform,the forward version was used to make up encoder and backward form is used to make up decoder. Note,this block architechture is a specially designed form of inception block.The details are as Fig.1.\par
Taking the advantage of indentity inception block as previously described,the symmetrical II-FCN is proposed.Ranging from the numbers of   the Identity Inception block,The II-FCN can be adapted to II-FCN8,II-FCN16,II-FCN32 and so on.Our experiments are all based on II-FCN32.We build bridges between forward Identity Inception block and backward Identity Inception block to prevent loss of texture and stable training.The architechture is specified in Fig.2.
\subsection{Loss Normalization}
Although our dynamic dataset augmentation increased image diversity greatly,the random crop and zooming methods would cause more unbalanced sample.Here,we adopted a reweighting or normalizing method.This makes our segmentation boundary more distinguishable.
Though our normalizing technichs,the loss curve could drop from random guessing value of binary classification,which can be used to evaluate whether the model has worked in early time.We formulated our methods as follows.\par
For a segmentation image \(s\),we write the pixel number of object as \(n\),we created a corresponding reweigthing filter \(f\).\par
The N can be computed through
\begin{equation}
n=\frac{\sum_{i=1}^{h}\sum_{j=1}^{w}(s[i][j])}{255}
\end{equation}\par
The percentage of object p is represented as
\begin{equation}
p=\frac{n}{w\times h}
\end{equation}
where \(w\) and \(h\) is the image width and height respectively.\par
Then we filled the corresponding percent of object and background in filter\(f\) as follows
\begin{equation}
       f[i][j] = 
        \begin{cases}
            p& \text{if $s[i][j] = 0$} \\
            1-p& \text{otherwise} \\
        \end{cases}
\end{equation}\par
At last,we normalized our filter by
\begin{equation}
f[i][j]=\frac{f[i][j]}{2\times p\times(1-p)}
\end{equation}\par
Our filter was used to normalize the final loss and reweigtht the gradients of back propagation,thus achived balanced training between melanoma and background.\par
\subsection{Jaccard Index Loss}
In this challenge,the evaluation metrics including sensitivity,specificity,pixel accuracy,dice coefficient,and jaccard index.The jaccard index is the final metrics to test the model.We put forward a method to approximate the jaccard index as a loss in our network.We specifies it at bellow.\par
We took softmax output of an image as \(A\)with probability of being the melanoma,the label as \(B\) with value 1 represent object and 0 represent background.First we computed the corresponding category number
\begin{equation}
	A = 
        \begin{cases}
            \frac{\max(A-0.5,0)}{A-0.5}& \text{if $A \neq 0.5$} \\
            1 & \text{otherwise} \\
        \end{cases}
\end{equation}
this maps the probability over 0.5 to label 1 and below 0.5 to label 0.\par
Then we approximate the jaccard index as the following
\begin{equation}
	J = 
        \begin{cases}
            \lg(\frac{\sum A+\sum B}{k\times(\sum A+\sum B)-\sum |A-B|}+(1-\frac{1}{k}))& \text{if $A \neq 0 and B \neq 0$} \\
            0 & \text{otherwise} \\
        \end{cases}
\end{equation}
where we took k as 1.1,then our jaccard index loss is between 0 and 1.004,which is comparable with the binary entropy loss.Then we took entropy loss with jaccard index loss together as our final optimization function.

\section{experiments and results}
\subsection{Experimental Setup}
Transfer learning is proven to be an efficent method to ease the training process of deep convolution models,which implies that the feature distribution of a large dataset may covers a small dataset more or less.However,we were going to train from scratch so as to prove the effictiveness of applying a bunch of image augmention strategies with our II-FCN model.\par
We appended the multi-scale trainning method apart from our dynamic dataset augmentation.we started from small scale training with image height and width of 252 and 380,then trainning at middle scale with image height and width of 444 and 688,finally,trainning at large scale with image height and width of 636 and 956.We set the batch size of training at different scale at 8,4,2,respectively.We used Adam optimizer in all scale with same setting of learning rate=0.0001,\(\beta_{1}=0.9\),,\(\beta_{2}=0.9999\),\(\epsilon=1e-8\).We performed our experiments with a 8GB nvidia GPU k5200.Our model is based on TensorFlow.
\subsection{Results}
We split 200 images from labeled data as validation set.The Fig.3 shows the jaccard  index  metrics from epoch 5 to 25.We masked from the softmax output with a probability large than 0.8 as a baseline.Then we performed conditional random field method to rectify the sgmentation.We found that after 10 epoch the jaccard index was above 0.8, the CRF method improved the baseline by a mean of about 0.005.\par
The figure also shows that our model kept learning from epoch to epoch,this demonstrated the effectiveness of our dataset augmentaion method.
\begin{figure}[htb]

\begin{minipage}[htb]{1.0\linewidth}
  \centering
  \centerline{\includegraphics[width=8.5cm,height=6.5cm]{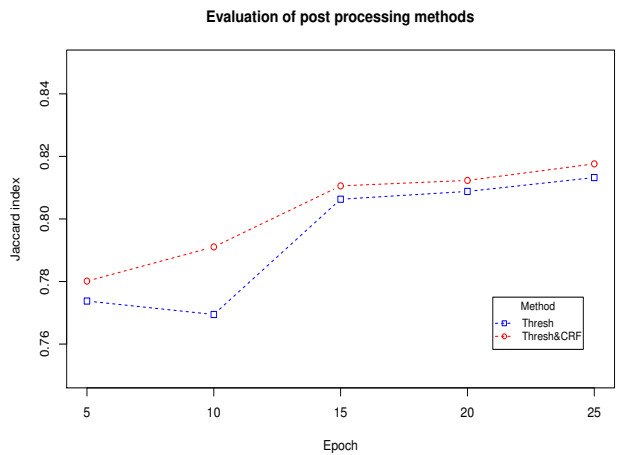}}
  \caption{Comparition of post processing methods}
   \label{fig:model}
\end{minipage}
\end{figure}
\section{Concludes}
We proposed an II-FCN model for melanoma segmentation with several training methods,including augmentaion,loss mormalization,jaccard index loss optimizing,dilation convolution and CRF processing.Our results showed that our model can automatically segment melanoma at high jaccard index without overfitting.

\bibliographystyle{IEEEbib}
\bibliography{strings,refs}

\end{document}